\documentclass[letterpaper, 10 pt, conference]{ieeeconf}  

\IEEEoverridecommandlockouts                              

\usepackage{cite}
\usepackage{amsmath,amssymb,amsfonts}
\usepackage{algorithmic}
\usepackage{algorithm}
\usepackage{graphicx}
\usepackage{textcomp}
\usepackage{xcolor}
\usepackage{booktabs}
\usepackage{longtable}
\usepackage{booktabs}   
\usepackage{multirow}   
\usepackage{graphicx}   
\usepackage{tabularx}
\usepackage{stfloats}   
\usepackage{algorithm}
\usepackage{tikz}
\usetikzlibrary{positioning} 

\usepackage{algorithmic}
\usepackage{subcaption}
\usepackage[utf8]{inputenc}
\usepackage{amsmath,amssymb}
\usepackage{background}
\usepackage{amsmath,bm}

\title{\LARGE \bf
Driving Beyond Privilege: Distilling Dense-Reward Knowledge into Sparse-Reward Policies}

\author{Feeza Khan Khanzada$^{1}$ and Jaerock Kwon$^{2}$
\thanks{$^{1}$Feeza Khan Khanzada, and $^{2}$Jaerock Kwon are with the Department of Electrical and Computer Engineering, University of Michigan-Dearborn,
        4901 Evergreen Rd, Dearborn, MI 48128, United States.
        {\tt\small \{feezakk, jrkwon\}@umich.edu}}%
}

\begin{document}

\maketitle
\thispagestyle{empty}
\pagestyle{empty}

\backgroundsetup{
  scale=1,
  angle=0,                      
  placement=left,               
  hshift=-10.5cm,
  color=black,
  opacity=1,                    
  contents={%
    \rotatebox{90}{%
      \parbox{23cm}{
        \small 
        Copyright may be transferred without notice, after which this version may no longer be accessible.
      }%
    }%
  },
}

\begin{abstract}
    We study how to exploit dense simulator rewards in vision-based autonomous driving without inheriting their misalignment with deployment metrics. In realistic simulators such as CARLA, privileged state (e.g., lane geometry, infractions, time-to-collision) can be converted into dense rewards that greatly stabilize and accelerate model-based reinforcement learning, but policies trained directly on these signals often overfit and fail to generalize when evaluated on sparse objectives such as route completion and collision-free overtaking. We propose reward-privileged world model distillation, a two-stage framework in which a teacher DreamerV3-style agent is first trained with a dense privileged reward, and only its latent dynamics are subsequently distilled into a student trained solely on sparse task rewards. Teacher and student share the same observation space (semantic bird’s-eye-view images); privileged information enters only through the teacher’s reward, and the student does not imitate the teacher’s actions or value estimates. Instead, the student’s world model is regularized to match the teacher’s latent dynamics while its policy is learned from scratch on sparse success/failure signals. In CARLA lane-following and overtaking benchmarks, sparse-reward students consistently outperform both dense-reward teachers and sparse-from-scratch baselines. On unseen lane-following routes, reward-privileged distillation improves success by approximately 23\% relative to the dense teacher while maintaining comparable or better safety. On overtaking, students retain near-perfect performance on training routes and achieve up to a 27× relative improvement in success on unseen routes, with improved lane keeping. These results show that dense rewards can be safely leveraged to learn richer dynamics models while keeping the deployed policy optimized strictly for sparse, deployment-aligned objectives.
\end{abstract}

\section{introduction}

Autonomous driving from raw sensory observations remains a central challenge in reinforcement learning (RL). Agents must reason over long horizons, operate under partial observability, and generalize to unseen road configurations, all while learning from sparse task feedback such as successfully completing the route without collision. Model-based RL with learned world models has emerged as a promising direction for such settings: by learning a latent dynamical model of the environment, agents can plan or imagine trajectories, improving sample efficiency and long-horizon credit assignment compared to model-free methods \cite{ha_world_2018, hafner_learning_2019, hafner_dream_2020}.

However, applying world models to realistic driving tasks still faces two practical obstacles: (i) sparse and delayed rewards in realistic objectives (e.g., success rate, safety violations) make policy learning unstable, and (ii) reward shaping with dense, privileged signals can improve training but often misaligns with the true evaluation criteria, hurting generalization \cite{vasan_revisiting_2024, hu_privileged_2024, nguyen_leveraging_2022, shenfeld_tgrl_2024, noauthor_specification_nodate, chaudhari_rlhf_2024}.

Simulation offers an attractive but underexploited lever in this context. Modern driving simulators expose privileged information such as precise ego and obstacle states, lane geometry, time-to-collision, or rule violations, which can be converted into dense reward signals \cite{rong_lgsvl_2020, shah_airsim_2017, dosovitskiy_carla_nodate}. These dense rewards are much easier to optimize than sparse success indicators and can greatly accelerate learning. However, policies trained directly on shaped rewards frequently exploit idiosyncrasies of the shaping function, leading to behaviors that score well on the dense proxy but perform poorly when evaluated on task-level metrics or under distribution shift (e.g., unseen routes or traffic configurations). This tension between reward shaping for learnability and reward fidelity for generalization is particularly acute in safety-critical domains such as lane following and overtaking in urban traffic.

Recent work has explored several ways to leverage privileged information within world-model RL. TWIST trains a teacher world model on privileged state observations and distills its latent dynamics into a student model that operates on images, enabling efficient sim-to-real transfer \cite{yamada_twist_2023}. Raw2Drive introduces a dual-stream architecture with separate world models for privileged and raw inputs, aligning them via a guidance mechanism to make end-to-end RL feasible in CARLA leaderboard settings \cite{yang_raw2drive_2025, noauthor_carla-simulatorleaderboard_2025, dosovitskiy_carla_nodate}. PIGDreamer and related approaches use privileged sensing to guide representation learning and safe control in partially observable tasks \cite{huang_pigdreamer_2025}. In parallel, a growing literature studies knowledge distillation in RL \cite{yi_survey_2025, hafez_continual_2024}, distilling policies \cite{rusu_policy_2016, czarnecki_distilling_nodate, sun_real-time_nodate}, value functions \cite{lyle_learning_nodate, amani_provably_2022, vincent_eau_2025}, or reward models from teacher to student to improve sample efficiency and stability \cite{fisch_robust_2025, hasanaliyev_unified_2025, noauthor_joint_nodate, chaudhary_novelty_nodate}, including in autonomous driving decision-making \cite{zhao_knowledge_2025, zhou_knowledge_2024, yu_distilldrive_2025, hong_knowledge_2024}. 

Despite this progress, existing methods typically assume that the teacher and student share the same task objective: either they both optimize the same reward, or they implicitly share a behavior distribution (e.g., via imitation). Privileged signals are usually encoded as additional observations (state vs. images) or as auxiliary losses, and the distilled student continues to optimize the same shaped or implicit objective as the teacher. In contrast, practical autonomous driving pipelines often rely on mismatched objectives: dense, simulator-defined rewards are used during training, while deployment is judged by sparse metrics such as route completion, collisions, infractions, and comfort. How to exploit privileged dense rewards only for representation learning, while ensuring that the final policy is optimized against the correct sparse objective, remains underexplored.

In this work, we investigate a simple but, to our knowledge, previously unexplored combination: distilling only the world‑model dynamics from a teacher trained with dense, simulator‑defined rewards into a student trained solely on sparse task rewards, without policy imitation or reward sharing. Concretely, we consider a model-based RL setting where both teacher and student observe the same sensor stream (e.g., RGB images or bird’s-eye-view representations). The teacher is trained using a privileged dense reward, computed from simulator state (e.g., lane deviation, time-to-collision, progress bonuses), and learns a latent world model together with a control policy. The student, in contrast, is trained only with a sparse task reward (e.g., success/failure of lane-following or overtaking episodes) and does not imitate the teacher’s policy. Instead, the student’s world model is regularized by a distillation loss that encourages its latent dynamics to match those of the teacher, while its policy is learned from scratch to maximize the sparse reward. Our central question is: can we leverage dense, simulator‑defined rewards exclusively to learn better dynamics models, while ensuring that the deployed policy is optimized only for the true sparse task objective?

We refer to this approach as reward‑privileged world‑model distillation. We use the term reward‑privileged to emphasize that privileged simulator information enters only through the teacher’s reward function, not through additional observations or features. Unlike prior world-model distillation work, which primarily transfers from privileged observations to realistic ones for the same reward \cite{yamada_twist_2023, huang_pigdreamer_2025, yang_raw2drive_2025}, our setting keeps the observation space fixed and introduces asymmetry solely through the reward signal. This design explicitly decouples the roles of privileged information: dense rewards are used to learn a rich, predictive world model in the teacher, but the control objective of the student remains the true sparse reward. Intuitively, the student inherits an improved model of environment dynamics without inheriting the teacher’s biases induced by reward shaping.

We instantiate this idea in a high-fidelity urban driving simulator and study two canonical tasks: lane following and overtaking in mixed traffic. Both tasks are evaluated using sparse, episode-level rewards that reflect realistic success metrics. Our results show that, on unseen routes within the same town, students trained with reward‑privileged world‑model distillation achieve substantially higher overtaking success rates—on the order of 27 times relative improvement—compared to their dense‑reward teachers, despite having access only to sparse task feedback during policy learning. At the same time, they outperform or match students trained from scratch with sparse rewards and no distillation. These findings suggest that dense rewards can be safely exploited for representation learning without baking their misspecifications into the final policy.

To place these observations on solid ground, we develop a systematic empirical study that compares reward‑privileged world‑model distillation to alternative ways of using privileged information and knowledge distillation in autonomous driving. We contrast world‑model‑only distillation with reward shaping, and analyse their impact on data efficiency and generalization across routes and traffic conditions. Our experiments indicate that transferring environment dynamics, rather than actions or value estimates, is particularly beneficial when the teacher’s reward is richer but imperfectly aligned with the true objective.

In summary, this paper makes the following contributions:

\begin{itemize}
    \item We formalize reward‑privileged world‑model distillation, in which a teacher world model is trained with a dense, simulator-defined reward, and only its latent dynamics are distilled into a student that learns its policy from sparse task rewards.
    \item We propose a concrete training procedure that combines standard model-based RL (DreamerV3-style latent imagination) with a distillation objective between teacher and student world models, while keeping the student’s policy optimization purely sparse-reward driven \cite{hafner_dream_2020}.
    \item In a realistic urban driving simulator, we demonstrate that our approach improves success rates and generalization in lane following and overtaking tasks compared to (i) dense‑reward teachers without distillation, (ii) Goal-conditioned sparse reward Hindsight Experience Replay based DreamerV3 model, and (iii) sparse‑reward students trained from scratch, with relative gains of up to 40–60\% on unseen overtaking routes.
    \item Through ablations and diagnostics, we investigate the role of world‑model distillation versus policy distillation and weight initialization, and show that reward‑privileged world‑model transfer can mitigate typical reward‑shaping pathologies while retaining the sample efficiency benefits of dense feedback.
\end{itemize}

By explicitly separating where privileged information enters (world‑model learning) from what objective the final policy optimizes (sparse task reward), our work offers a simple and practical tool for deploying world‑model RL in autonomous driving scenarios where dense reward design is convenient in simulation but misaligned with real‑world evaluation criteria.

\section{Related Work}


\paragraph{Model-based RL for autonomous driving.}
Model-based reinforcement learning (MBRL) has become increasingly prominent in autonomous driving by learning world models that predict environment dynamics for planning or policy optimization \cite{ha_world_2018, hafner_learning_2019, hafner_dream_2020}. Planning in a learned latent model (as in Dreamer) can significantly improve sample efficiency and long-horizon credit assignment compared to model-free methods. Several works extend world-model RL to realistic driving scenarios using simulators like CARLA and nuPlan \cite{noauthor_nuplan_nodate} with camera, LiDAR, or bird’s-eye-view (BEV) inputs. Early systems such as PlanT and Roach applied MBRL to CARLA but relied on complex, multi-term reward functions and often required privileged inputs, which limited scalability and robustness \cite{noauthor_plant_nodate, noauthor_end--end_nodate}. 

More recent work focuses on scaling world models and simplifying objectives. Think2Drive applies a DreamerV3-style \cite{hafner_mastering_2024} latent dynamics model with actor–critic learning to urban driving tasks, showing that long-horizon planning in a learned latent space can handle dense traffic scenarios \cite{li_think2drive_2024}. InDRiVE similarly adopts a Dreamer-style world model in CARLA, but trains it 
initially using purely intrinsic, disagreement-based rewards before few-shot 
fine-tuning on lane-following and collision-avoidance tasks, achieving strong 
zero-shot and few-shot generalization across towns \cite{khanzada_indrive_2025}. CaRL demonstrates that a simplified reward centered on route completion with infractions as termination or heavy penalties enables stable large-scale PPO \cite{schulman_proximal_2017} training of a privileged world-model planner, achieving state-of-the-art driving scores on CARLA and nuPlan while outperforming prior RL methods based on finely tuned reward shaping \cite{benjamins_carl_2021}. These results suggest that streamlined reward designs can avoid poor local optima and are amenable to large-scale training.

Several works integrate world models with realistic perception and privileged information. Raw2Drive trains two parallel world models: a privileged model that uses low-dimensional simulator state (maps, object locations) and a raw-sensor model using images \cite{yang_raw2drive_2025}. A guidance mechanism aligns their latent predictions, effectively distilling knowledge from the easy state space into the harder vision model; the resulting policy is the first pure-RL agent to achieve state-of-the-art performance on the CARLA Leaderboard 2.0. AdaWM pre-trains a world model on offline driving logs and then adapts it online by selectively updating either the model or the policy when prediction errors indicate distribution shift, mitigating catastrophic degradation during transfer \cite{wang_adawm_2025}. 

Another line blends offline data and world models. World on Rails learns a global world model from pre-recorded driving logs (with access to privileged log information such as maps and other vehicles’ trajectories), performs dynamic programming in this model to compute action-values, and uses these values to supervise a reactive image-only policy \cite{chen_learning_2021}. The resulting policy, which never sees privileged inputs at execution time, outperforms both imitation learning and standard RL on CARLA benchmarks. Overall, these works show that world models, often aided by privileged inputs, are powerful tools for autonomous driving. However, the teacher and student (or planner and policy) typically share the same underlying reward or driving score. Complementary evidence from off-road settings indicates that seemingly low-level 
perception design choices such as camera region-of-interest and dataset size can 
strongly affect the performance and robustness of end-to-end controllers 
\cite{khanzada2024analytical}, underscoring the importance of representation 
choices alongside the RL algorithm.


\paragraph{Knowledge distillation and privileged information.}
Knowledge distillation is widely used in robotics and autonomous driving to transfer information from a teacher with privileged access or easier inputs to a student constrained to realistic observations. Learning by Cheating (LBC) trains a privileged “cheater” agent that sees full simulator state (BEV layout, all actors) and then distills its behavior into a purely vision-based student via supervised imitation \cite{chen_learning_2019}. This two-stage pipeline achieves near-perfect success on CARLA by first solving an easier privileged problem, then transferring the solution to raw images. 

Beyond policy distillation, several works explicitly distill world models or representations. TWIST trains a teacher world model on low-dimensional simulator state and a student model on images, using paired state–image data to distill the teacher’s latent dynamics and representation into the student \cite{yamada_twist_2023}. This improves sim-to-real transfer for vision-based MBRL. Raw2Drive can similarly be viewed as world-model distillation: the privileged model’s dynamics and cost predictions act as targets for the raw-sensor model during guided rollouts, transferring planning competence from state to pixels \cite{yang_raw2drive_2025}. World on Rails uses a learned world model as an offline planner and distills its action-values into a reactive policy \cite{chen_learning_2021}. 

More generally, policy distillation and representation transfer are used to compress multiple experts into a single student \cite{rusu_policy_2016 , teh_distral_2017} and to inject knowledge from large pre-trained vision or vision–language models into RL policies. These works follow a common pattern: a teacher solves an easier or more informed version of the task (via privileged state, offline planning, or larger models), and a student learns to act under realistic constraints. Our work fits this teacher–student paradigm but differs in two respects: (i) teacher and student share the same observation space, and (ii) we distill only the \emph{world model} of a dense-reward teacher, while the student’s policy is trained exclusively on a sparse task reward and never imitates the teacher’s actions or value estimates.


\paragraph{Privileged information and rewards under partial observability.}
A complementary line of work studies how to exploit privileged information during training while keeping the deployment interface fixed. Asymmetric actor–critic methods allow the critic to access privileged information (e.g., full state or goal location) while the actor sees only partial observations, and show that policy gradients remain unbiased even when the critic receives arbitrary extra signals \cite{pinto_asymmetric_2017}. This significantly improves learning in partially observed navigation and multi-agent tasks, and is now a standard pattern (centralized critic, decentralized policies). Privileged information dropout (PID) instead injects privileged features directly into the observation during training but randomly masks them to avoid over-reliance, improving sample efficiency in partially observed domains without requiring a separate teacher \cite{kamienny_privileged_2020}. 

Sensory scaffolding approaches such as Scaffolder provide extra sensor modalities (e.g., LiDAR, pose, object trackers) at training time to guide representation learning in policies, world models, and exploration objectives \cite{hu_privileged_2024}. The final policies run without these privileged streams, but benefit from the structure induced during training; the same privileged sensors can also be used to define more informative rewards. Other common strategies in sparse-reward environments include using hindsight information (as in Hindsight Experience Replay \cite{andrychowicz_hindsight_2018}) or simulator-based distance-to-goal signals as shaping rewards during training, even when such quantities are not available at deployment \cite{trott_keeping_2019, madhavan_role_2022}. These works demonstrate that privileged observations and rewards can be safely used to scaffold learning as long as the deployed policy does not depend on them directly.

Our setting is orthogonal in the way that the teacher and student observe exactly the same sensor stream, and the asymmetry comes solely through the \emph{reward}. The teacher is trained with a dense, simulator-defined reward, whereas the student only ever optimizes the sparse task reward. Privileged information enters only through the teacher’s learning signal and the induced world model.


\paragraph{Dense vs.\ sparse rewards and generalization.}
The impact of dense versus sparse rewards on learning and generalization has been revisited in several recent studies. Dense rewards typically accelerate early training by providing frequent feedback and reducing the credit-assignment burden, but can induce reward hacking and misalignment with the true task objective \cite{amodei_concrete_2016}. Vasan et al.\ show that while dense shaping speeds up learning in goal-reaching tasks, policies trained with purely sparse minimum-time rewards often achieve higher final success rates and better trajectories; dense-reward agents tend to overfit to the hand-crafted shaping terms \cite{vasan_revisiting_2024}. Booth et al.\ analyze the “perils of trial-and-error reward design” and show how iterating on complex multi-term rewards can cause overfitting to specific environments and algorithms, degrading generalization \cite{booth_perils_2023}. 

Similar phenomena appear in autonomous driving. Early RL driving work combined lane keeping, speed, comfort, and collision penalties in heavily engineered reward functions, which required delicate tuning and often produced conservative or odd behaviors \cite{noauthor_end--end_nodate, delavari_comprehensive_2025}. CaRL demonstrated that simplifying the reward to essentially route completion with infractions as hard penalties yields better policies and more scalable optimization than complex shaped rewards, and that the resulting policy can even perform better on dense driving-score metrics than agents trained directly on those shaped scores \cite{benjamins_carl_2021}. These results reinforce the view that dense or privileged rewards are powerful training tools but can compromise generalization if used as the primary policy objective.

Our work takes this lesson explicitly: we use dense, simulator-defined rewards only to train a teacher world model, and then distill its latent dynamics into a student whose policy is optimized solely for the sparse task reward. Reward-privileged world-model distillation thus aims to inherit the representation and planning benefits of dense rewards while avoiding the behavioral biases that shaped rewards can introduce into the deployed policy.

\section{Method}

\begin{figure*}[t]
  \centering
  \includegraphics[width=\linewidth]{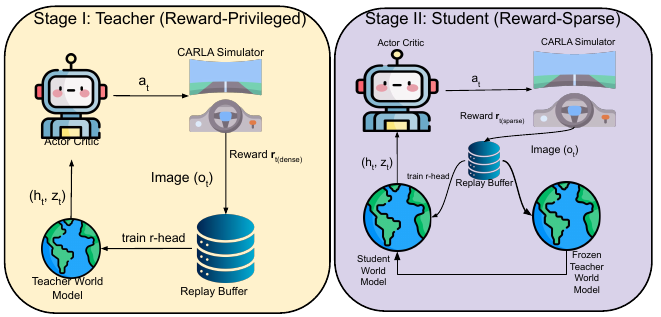}
  \caption{Two–stage training pipeline. 
  \textbf{Left:} Reward‑privileged teacher. An actor–critic interacts with the CARLA simulator using actions $a_t$, receives images $o_t$ and dense reward $r_t^{\text{dense}}$, and stores transitions in a replay buffer. A world model is trained on this buffer and provides latent states $(h_t, z_t)$ to the actor–critic.  
  \textbf{Right:} Reward‑sparse student. A second actor–critic interacts with CARLA using only sparse reward $r_t^{\text{sparse}}$. The replay buffer is shared between a student world model and a frozen copy of the teacher world model; the student is regularized to match the teacher’s latents, while its own latents drive policy learning.}
  \label{fig:teacher-student}
\end{figure*}

We aim to exploit privileged dense rewards available in simulation to improve world-model learning, while ensuring that the final control policy is optimized only for a sparse, task-level objective. To this end, we propose reward-privileged world-model distillation, a teacher–student framework in which a teacher model-based RL agent is trained with a dense simulator-defined reward, and only its world model is distilled into a sparse-reward student. The student uses the distilled world model to learn a policy from scratch under the sparse task reward, without imitating the teacher's actions or value estimates. Fig. \ref{fig:teacher-student} represents a high level illustration of the methodology described in this section.

\subsection{Problem formulation}

We consider a partially observable Markov decision process (POMDP)
\[
\mathcal{M} = (\mathcal{S}, \mathcal{A}, \mathcal{O}, p, r^{\text{sparse}}, r^{\text{dense}}, \gamma),
\]
where $\mathcal{S}$ is the set of environment states, $\mathcal{A}$ the action space (discrete vehicle controls), and $\mathcal{O}$ the observation space (a semantic segmented Bird-Eye View (BEV) image). The transition kernel $p(s_{t+1} \mid s_t, a_t)$ is unknown. The simulator exposes two reward functions:

- a \emph{sparse} task reward $r^{\text{sparse}}(s_t, a_t)$ encoding episode-level success metrics (e.g.\ route completion without collisions or major infractions), and  
- a \emph{dense}, privileged reward $r^{\text{dense}}(s_t, a_t)$ constructed from simulator state (e.g.\ lane deviation, time-to-collision, progress bonuses).

At deployment, only $r^{\text{sparse}}$ is meaningful; $r^{\text{dense}}$ is purely a training-time artefact and may be misaligned with the true objective. Both teacher and student observe the same observation process
\[
o_t \sim p(o_t \mid s_t),
\]
and must output actions according to a stochastic policy $\pi(a_t \mid o_{\le t})$.

The student’s goal is to maximize the expected discounted return under the sparse reward:
\[
J_{\text{student}}(\pi) = \mathbb{E}\Big[ \sum_{t=0}^{T} \gamma^t \, r^{\text{sparse}}(s_t, a_t) \Big].
\]
The teacher instead maximizes the return under $r^{\text{dense}}$.

\subsection{World-model RL backbone}

Both teacher and student are built on the same world-model RL backbone, following Dreamer-style latent dynamics models \cite{hafner_dream_2020}. We summarize the main components here.

We introduce a latent state space $\mathcal{z}$ and parameterize a recurrent state-space model (RSSM) with parameters $\theta$:
\begin{align}
p_\theta(z_0) &= \mathcal{N}(0, I), \\
p_\theta(z_t \mid z_{t-1}, a_{t-1}) &= \text{prior transition}, \\
q_\theta(z_t \mid z_{t-1}, a_{t-1}, o_t) &= \text{posterior encoder}, \\
p_\theta(o_t \mid z_t) &= \text{observation decoder}, \\
p_\theta(r \mid z_t) &= \text{reward head}, \\
p_\theta(v \mid z_t) &= \text{value head}.
\end{align}

Given a sequence $(o_{0:T}, a_{0:T-1})$, the model is trained to maximize an evidence lower bound (ELBO) that reconstructs observations and rewards and regularizes the posterior toward the prior:
\begin{equation}
\begin{aligned}
\mathcal{L}_{\text{wm}}(\theta)
    = \sum_{t=0}^T \Big[
        &\mathbb{E}_{q_\theta}[\log p_\theta(o_t \mid z_t)] \\
        &+ \mathbb{E}_{q_\theta}[\log p_\theta(r_t \mid z_t, a_t)] \\
        &- \beta\, \mathrm{KL}\big(q_\theta(z_t\mid\cdot) \,\|\, p_\theta(z_t\mid\cdot)\big)
    \Big].
\end{aligned}
\end{equation}

An actor–critic is trained on imagined trajectories generated by unrolling the learned latent dynamics model. Starting from a posterior sample $z_t \sim q_\theta(\cdot)$, we roll out for $H$ steps:
\begin{align}
z_{t+1} &\sim p_\theta(z_{t+1} \mid z_t, a_t), \\
a_t &\sim \pi_\phi(a_t \mid z_t), \\
\hat{r}_t &\sim p_\theta(r \mid z_t, a_t),
\end{align}
and optimize an actor $\pi_\phi$ and critic $v_\psi$ to maximize the imagined return. We use standard Dreamer-style objectives (entropy-regularized actor loss and temporal-difference critic loss) and omit details for brevity.

\subsection{Reward-privileged teacher training}

The \emph{teacher} agent $(\theta_T, \phi_T, \psi_T)$ is trained in the simulator using the dense reward $r^{\text{dense}}$ as its learning signal. The teacher’s world model includes a reward head $p_{\theta_T}(r^{\text{dense}} \mid z_t, a_t)$, and the RL objective is
\[
J_{\text{teacher}} = \mathbb{E}\Big[ \sum_{t=0}^T \gamma^t r^{\text{dense}}(s_t, a_t) \Big].
\]

Training alternates between:

1. collecting trajectories $(o_t, a_t, r^{\text{dense}}_t)$ by interacting with the environment under $\pi_{\phi_T}$,  
2. updating the world model parameters $\theta_T$ by maximizing the ELBO with reconstruction of $o_t$ and $r^{\text{dense}}_t$, and  
3. updating the actor–critic $(\phi_T, \psi_T)$ from imagined rollouts in the teacher’s latent space using $p_{\theta_T}(r^{\text{dense}} \mid z_t)$ as the reward model.

Training continues until the teacher converges on the dense-reward objective. After convergence, we \emph{freeze} all teacher parameters. At this point, the teacher world model encodes a latent representation $z_t^T$ and transition dynamics that are shaped by the dense reward, but its policy may be suboptimal or misaligned with respect to $r^{\text{sparse}}$.

\subsection{Sparse-reward student with world-model distillation}

The \emph{student} agent $(\theta_S, \phi_S, \psi_S)$ is trained only with the sparse task reward $r^{\text{sparse}}$. The student uses the same world-model architecture and actor–critic structure as the teacher, but with two key differences:

1. the reward head predicts $r^{\text{sparse}}$ instead of $r^{\text{dense}}$, and  
2. the world model is regularized by a distillation loss that aligns its latent dynamics with those of the frozen teacher.

\paragraph{Latent dynamics alignment.}
Given batches of trajectories from the replay buffer, we obtain latent sequences from both teacher and student posteriors:
\begin{align}
z_t^T &\sim q_{\theta_T}(z_t \mid z_{t-1}^T, a_{t-1}, o_t), \\
z_t^S &\sim q_{\theta_S}(z_t \mid z_{t-1}^S, a_{t-1}, o_t).
\end{align}




We define a latent alignment loss that encourages the student’s posterior to match the teacher’s. We align both the deterministic and stochastic parts of the latent state: we penalize the squared distance between teacher and student deterministic latents, and a KL divergence between their stochastic latent distributions. Denoting by $h_t^\text{det}$ the deterministic latent and by $q_t^\text{stoch}$ the stochastic latent distribution, we minimize


\begin{equation}
\begin{aligned}
\mathcal{L}_{\text{dist}}(\theta_S) 
  &= \frac{1}{T} \sum_{t=0}^T 
     \Big(
       \lambda_\text{det} \big\| h_{t,S}^\text{det} - h_{t,T}^\text{det} \big\|_2^2 \\
  &\quad + \lambda_\text{stoch}\, 
       \mathrm{KL}\big(q_{t,T}^\text{stoch} \,\|\, q_{t,S}^\text{stoch}\big)
     \Big),
\end{aligned}
\end{equation}

where $\lambda_\text{det}, \lambda_\text{stoch}$ are fixed loss weights.
The teacher’s parameters are fixed, so gradients flow only into the student. This loss encourages the student to encode observations into a latent space compatible with the teacher’s dense-reward-shaped representation, while still being free to adapt where needed for the sparse objective.




\begin{equation}
\begin{aligned}
\mathcal{L}_{\text{trans}}(\theta_S) 
  &= \frac{1}{T} \sum_{t=1}^T 
     \mathrm{KL}\!\big(
       p_{\theta_T}(z_t \mid z_{t-1}^T, a_{t-1}) \\
  &\qquad\qquad\qquad
       \,\big\|\,
       p_{\theta_S}(z_t \mid z_{t-1}^S, a_{t-1})
     \big),
\end{aligned}
\end{equation}

\paragraph{Student world-model and RL objective.}
The student’s world-model loss combines standard reconstruction of observations and sparse rewards with the distillation term:



\begin{equation}
\begin{split}
\mathcal{L}_{\text{wm-student}}(\theta_S)
  &= \mathcal{L}_{\text{recon}}^{\text{sparse}}(\theta_S) \\
  &\quad + \lambda_{\text{post}} \mathcal{L}_{\text{dist}}(\theta_S)
         + \lambda_{\text{prior}} \mathcal{L}_{\text{trans}}(\theta_S),
\end{split}
\label{eq:wm_student}
\end{equation}

where $\mathcal{L}_{\text{recon}}^{\text{sparse}}$ is the ELBO-style loss using $r^{\text{sparse}}$ instead of $r^{\text{dense}}$, and $\lambda_{\text{dist}}$ controls the influence of distillation.

Crucially, the student’s actor and critic are trained \emph{only} on the sparse reward model $p_{\theta_S}(r^{\text{sparse}} \mid z_t)$. Imagined rollouts are generated using the student dynamics $p_{\theta_S}(z_{t+1} \mid z_t, a_t)$, and the actor–critic is optimized to maximize the expected sparse-return in imagination:

\begin{equation}
\begin{aligned}
J_{\text{student}}(\phi_S)
&= \mathbb{E}\Big[ \sum_{k=0}^{H-1} \gamma^k \hat{r}^{\text{sparse}}_{t+k} \Big], \\
\hat{r}^{\text{sparse}}_{t+k}
&\sim p_{\theta_S}(\cdot \mid z_{t+k}).
\end{aligned}
\end{equation}

There is deliberately \emph{no} policy or value distillation term from the teacher to the student. Any difference between teacher and student behavior is mediated solely through the world model; the student is free to discover a policy that improves the true sparse objective, even if this deviates from the teacher’s dense-reward policy.

\subsection{Training procedure}

Training proceeds in two stages.

\paragraph{Stage 1: teacher training.}
We initialize the teacher’s world model and actor–critic randomly and train them on the dense reward $r^{\text{dense}}$ until convergence. During this stage, the simulator provides $r^{\text{dense}}$ and we do not use $r^{\text{sparse}}$ at all. The output is a frozen teacher $(\theta_T, \phi_T, \psi_T)$ and a replay buffer of trajectories $(o_t, a_t, r^{\text{dense}}_t)$.

\paragraph{Stage 2: student training with distillation.}
We initialize the student parameters $(\theta_S, \phi_S, \psi_S)$ independently (we evaluate teacher-initialization as a separate baseline). The environment now exposes only the sparse reward $r^{\text{sparse}}$. At each iteration we:

\begin{enumerate}
\item Collect trajectories $(o_t, a_t, r^{\text{sparse}}_t)$ by acting with the current student policy $\pi_{\phi_S}$.
\item For each trajectory, run the frozen teacher and the student encoders in parallel to obtain latent sequences $z_t^T$ and $z_t^S$.
\item Update the student world model parameters $\theta_S$ by minimizing $\mathcal{L}_{\text{wm-student}}$ in Eq.~\eqref{eq:wm_student}.
\item Update the student actor–critic $(\phi_S, \psi_S)$ using Dreamer-style imagined rollouts in the student’s latent space, with sparse rewards.
\end{enumerate}

Note that although the teacher was originally trained on dense rewards, its world model can be run on new trajectories without access to $r^{\text{dense}}$, since it conditions only on $(o_t, a_t)$. Hence we can compute distillation losses on trajectories that the teacher never saw and in regions where the teacher policy would not visit, which empirically helps student generalization.

\subsection{Discussion}

The proposed framework separates where privileged information enters from what objective the final policy optimizes. Dense, simulator-defined rewards are used exclusively during teacher training to shape the teacher world model (and its associated actor–critic), but only the world model is reused by the student.The student reuses this structured latent space via world-model distillation, while its policy is optimized only for the sparse task reward. This design aims to retain the representation and planning benefits of dense reward while avoiding inheriting its potential misspecification. In Section~\ref{sec:results}, we show that this yields higher success rates and better generalization on lane-following and overtaking tasks than (i) dense-reward teachers evaluated directly, (ii) sparse-reward students without distillation, and (iii) students with policy/value distillation or naive weight initialization.

\section{Experimental Setup} \label{sec:experiments}

\subsubsection{Lane Following}

\paragraph{Town and map.}
We use CARLA Town07, a compact rural--suburban map with winding roads, sparse buildings, and moderate elevation changes, designed for off‑highway and low‑traffic driving.

\paragraph{Routes and train/test split.}
We design a long reference route and partition it into a set of non‑overlapping training sub‑routes. At training time, one training sub‑route is sampled uniformly at the beginning of each episode.

For evaluation, we use a fixed set of routes in Town07 consisting of the multi‑turn, single‑turn, and straight routes. All methods are trained only on the training sub‑routes and evaluated on both training and evaluation routes under three random seeds.

\paragraph{Actors.}
Each episode contains a single ego vehicle controlled by the RL agent.
No additional dynamic traffic or pedestrians are spawned in this task; only static scenery from Town07 is present.

\paragraph{Goal.}
The ego must follow the assigned route and reach the end‑of‑route trigger without colliding or leaving the drivable lane.

\paragraph{Observations $o_t$.}
The ego receives:
\begin{itemize}
    \item a top‑down semantic‑segmentation view from a camera mounted at height $z{=}20\,\mathrm{m}$ with pitch $-90^\circ$, rendered with a Cityscapes‑style palette and downsampled to $128{\times}128$;
    \item binary collision and lane‑invasion flags.
\end{itemize}

\paragraph{Action space.}
We use a discrete grid of throttle and steering commands,
\[
\mathcal{A}
= \{0.0,\,0.3\} \times \{-0.2,\,-0.1,\,0,\,0.1,\,0.2\},
\]
yielding $|\mathcal{A}| = 10$ discrete actions.

\paragraph{Horizon and termination.}
Episodes are capped at $T=1000$ control steps (100\,s at 10\,Hz) unless terminated earlier.
Termination criteria include success, collision, lane departure, or time‑out, as detailed in Sec.~\ref{sec:term-criteria}.
Sensors include the semantic‑segmentation camera and CARLA's standard collision and lane‑invasion sensors.

\subsubsection{Overtaking}\label{sec:overtake-env}

\paragraph{Town and map.}
For overtaking, we use CARLA Town04, a compact European‑style urban centre with dense road networks, tight intersections, and varied elevation, including multi‑lane roads and overpasses.

\paragraph{Routes and train/test split.}
We construct a set of training overtaking routes and a disjoint set of unseen test routes in Town04. Each route specifies the spawn poses of the ego and lead vehicle and the terminal line.
All methods are trained only on the training routes and evaluated on both training and unseen routes under three random seeds.

\paragraph{Actors.}
Each episode involves a single ego vehicle and a scripted lead vehicle.
Spawn points are chosen from the predefined route configurations, and we iterate through a random permutation of these configurations across episodes.
The lead vehicle follows a lane‑keeping PID controller and, when the ego approaches within $20\,\mathrm{m}$, performs a small lateral swing manoeuvre
(with parameters as in our code: swing steer = 0.04, swing amplitude $0.2\,\mathrm{m}$).
No additional traffic or pedestrians are spawned.

\paragraph{Goal.}
The ego starts behind the lead vehicle, must overtake it, and must return to the original lane before crossing the terminal line without colliding or leaving the drivable lane.
We log a separate \emph{Overtaken} indicator when the ego passes the lead vehicle at least once during the episode, regardless of final success.

\paragraph{Observations $o_t$.}
The ego receives:
\begin{itemize}
    \item a top‑down semantic‑segmentation view from a camera mounted at height $z{=}20\,\mathrm{m}$ with pitch $-90^\circ$, rendered with a Cityscapes‑style palette and downsampled to $128{\times}128$;
    \item binary collision and lane‑invasion flags.
\end{itemize}
Teacher and student again share the same observation space.

\paragraph{Action space.}
We use the same discrete action grid as in lane following, with $|\mathcal{A}| = 10$.

\paragraph{Horizon and termination.}
Episodes are capped at $T=1000$ control steps (100\,s at 10\,Hz) and can terminate earlier by success, collision, lane departure, or time‑out, as specified in Sec.~\ref{sec:term-criteria}.

\subsection{Teacher vs.\ Student Rewards}\label{sec:rewards}
The \textbf{teacher} is trained on a \emph{dense}, simulator-defined reward $r_t^{\text{dense}}$ computed from privileged simulator state (lane geometry, time-to-collision, infractions, ground-truth progress). The \textbf{student} is trained on a \emph{sparse}, task-level reward $r_t^{\text{sparse}}$ reflecting deployment metrics. Neither agent ever receives privileged state as an observation.



\subsubsection{Lane Following: Teacher Dense Reward}
The teacher uses a dense shaped reward that combines:

\begin{itemize}
    \item a quadratic penalty on lateral distance to the lane centre,
    \item a quadratic penalty on heading error to the road tangent,
    \item a speed term that encourages forward velocity along the lane at approximately $20\,\mathrm{km/h}$ and discourages lateral motion,
    \item a gentle low-speed penalty for speeds below $1\,\mathrm{km/h}$ and a strong penalty after $10\,\mathrm{s}$ of stalling,
    \item per-event penalties for lane invasions and collisions, and
    \item a progress term proportional to the reduction in distance to the goal, with a terminal bonus of $+200$ on reaching the goal within $2\,\mathrm{m}$.
\end{itemize}
 
The overall dense reward $r_t^{\text{dense}}$ is the sum of these components.

\subsubsection{Lane Following: Student Sparse Reward}
Episode-level success only:

\begin{equation}
r_t^{\text{sparse}} =
\begin{cases}
1, &
\begin{aligned}
  &\text{episode ends and goal reached}\\
  &\text{within }2\,\mathrm{m}\text{ without collision},
\end{aligned}
\\[4pt]
0, & \text{otherwise}.
\end{cases}
\end{equation}

No per-step shaping is used for the student. The sparse reward is zero at all intermediate steps and equals 1 only on the terminal transition if the episode ends in a collision‑free goal within 2 m.

\subsubsection{Overtaking: Teacher Dense Reward}\label{sec:overtake-dense}
The environment computes a shaped reward that sums:
\begin{itemize}
    \item waypoint progress and speed alignment along the lane tangent (desired $5\,\mathrm{m/s}$),
    \item gentle low-speed penalties for speed $<1\,\mathrm{km/h}$ and a strong penalty after $10\,\mathrm{s}$,
    \item per-event collision penalties,
    \item lane-discipline shaping: encourage staying in lane until close to the lead vehicle and penalize early lane changes,
    \item event rewards for \emph{exceed} (ego passes the lead) and \emph{overtake} (pass and return to the original lane).
\end{itemize}

Scales match the released code (e.g., $+200$ for the first \emph{exceed} and \emph{overtake} events, $-500$ per collision).

\subsubsection{Overtaking: Student Sparse Reward}

Binary episode-level success:

\begin{equation}
r_t^{\text{sparse}} =
\begin{cases}
1, & \text{overtake success},\\
0, & \text{otherwise}.
\end{cases}
\end{equation}

\subsection{Termination Criteria}\label{sec:term-criteria}

Episodes terminate when any of the following conditions is met:

\begin{itemize}
    \item the goal is reached (distance to goal $<2\,\mathrm{m}$),
    \item a collision is detected,
    \item a time limit is exceeded ($T=1000$ control steps),
    \item the ego remains below $1\,\mathrm{km/h}$ for more than $10\,\mathrm{s}$,
    \item for overtaking, the ego leaves a predefined lateral band around the spawn lane centre, or
    \item the ego has driven past the goal line without satisfying the success conditions.
\end{itemize}

\subsection{Models and Training}

\subsubsection{World models}

Teacher and student share the same DreamerV3-style latent dynamics model \cite{hafner_mastering_2024}: a convolutional encoder, a recurrent state-space model (RSSM) with a high-dimensional deterministic state and discrete stochastic latents, and image/reward/continuation decoders. Architectures and optimizer hyperparameters are identical; they differ only in the reward signal (dense vs.\ sparse) and the additional distillation losses applied to the student.

\paragraph{Architecture.}
Input observations are $128{\times}128$ semantic BEV images. Images are encoded with a ResNet-style CNN (depth 96, stride-based downsampling to a $4{\times}4$ minimum resolution) followed by a 2-layer MLP (16 units, SiLU activations, layer norm). Inputs are symlog-preprocessed.

The RSSM maintains a 4096-dimensional deterministic state and 32 categorical stochastic variables with 32 classes each. It uses SiLU activations, layer norm, a learned initial state, and unimix 0.01. The image decoder mirrors the encoder and conditions on the concatenation of deterministic and stochastic latents. Reward and continuation heads are MLPs on the latent state.

Table~\ref{tab:wm-arch} summarizes the main world-model hyperparameters.

\begin{table}[t]
  \centering
  \caption{World model and representation hyperparameters.}
  \label{tab:wm-arch}
  \begin{tabularx}{\columnwidth}{@{}l@{\hspace{0.75em}}X@{}}
    \toprule
    \textbf{Component} & \textbf{Setting} \\
    \midrule
    Input & $128{\times}128$ semantic BEV image \\
    Encoder CNN & ResNet, depth 96, stride resize, min res $4{\times}4$ \\
    Encoder MLP & 2 layers, 16 units, SiLU, layer norm \\
    Latent state (deterministic) & 4096-dim \\
    Latent state (stochastic) & $32 \times 32$-class categorical variables \\
    RSSM details & SiLU, layer norm, learned init, unimix $0.01$ \\
    Image decoder & ResNet, depth 96, inputs: [deter, stoch] \\
    Reward head & 5 layers, 1024 units, symlog-disc (255 bins) \\
    Continuation head & 5 layers, 1024 units, Bernoulli \\
    World-model losses & image: 1.0, reward: 1.0, cont: 1.0, dyn: 0.5, rep: 0.1 \\
    \bottomrule
  \end{tabularx}
\end{table}

\subsubsection{Actor and critic.}
Actor and value networks operate on the RSSM latent $(h_t, z_t)$. Each is a 5-layer MLP (1024 units, SiLU, layer norm) with inputs [deter, stoch]. The actor outputs a one-hot discrete distribution over CARLA actions with unimix 0.01. The critic predicts a symlog-discretized value distribution with 255 bins. We use a V-function critic with a slow target network (update fraction 0.02), return $\lambda$-trace with $\lambda = 0.95$, and imagination horizon $H=15$.

\begin{table}[t]
  \centering
  \caption{Actor--critic and optimization hyperparameters.}
  \label{tab:ac-train}
  \begin{tabularx}{\columnwidth}{@{}l@{\hspace{0.75em}}X@{}}
    \toprule
    \textbf{Component} & \textbf{Setting} \\
    \midrule
    Actor network & 5 layers, 1024 units, SiLU, layer norm \\
    Actor dist (discrete) & one-hot, unimix 0.01 \\
    Critic network & 5 layers, 1024 units, SiLU, layer norm \\
    Critic dist & symlog-disc (255 bins) \\
    Imag. horizon $H$ & 15 steps \\
    Return $\lambda$ & 0.95 \\
    Discount horizon & 333 ($\gamma = 1 - 1/333$) \\
    World-model opt & Adam, lr $1\times10^{-4}$, $\epsilon=10^{-8}$ \\
    Actor opt & Adam, lr $3\times10^{-5}$, $\epsilon=10^{-5}$, clip 100 \\
    Critic opt & Adam, lr $3\times10^{-5}$, $\epsilon=10^{-5}$, clip 100 \\
    Entropy coeff. & $3\times10^{-4}$ \\
    \midrule
    Replay buffer & size $10^5$, prioritized ($\alpha=0.7$, $\beta=0.7$) \\
    Batch size / length & 16 / 64 \\
    Train:env ratio & 512 grad steps per env step \\
    Total steps & $10^5$ gradient updates \\
    Reward scale (teacher) & $\times 100$ on dense reward \\
    \bottomrule
  \end{tabularx}
\end{table}

\subsubsection{Distillation losses (student only).}
For the student, the world-model loss is augmented with latent-alignment terms:
\begin{itemize}
\item posterior alignment: MSE between teacher and student deterministic latents ($\lambda_{\text{post-deter}} = 1.0$) and KL between stochastic latents ($\lambda_{\text{post-stoch}} = 1.0$);
\item prior/transition alignment: MSE between deterministic priors ($\lambda_{\text{prior-deter}} = 1.0$) and KL between stochastic priors ($\lambda_{\text{prior-stoch}} = 1.0$).
\end{itemize}

Teacher and student therefore share architecture and optimizer settings; the only differences are the reward signal (dense vs.\ sparse) and the additional world-model distillation losses applied during student training.

We train in two phases. In Phase I, a teacher agent is trained on dense reward only.
In Phase II, we freeze the teacher and train a sparse-reward student with the same
optimizers and horizons, augmenting the student world-model loss with the
distillation terms described above.



\subsection{Baselines}

We compare the following variants, all using the same Dreamer-style backbone and hyperparameters.

\begin{itemize}
    \item \textbf{Dense (DreamerV3):}
    A standard Dreamer agent trained and evaluated on the dense simulator reward $r^{\text{dense}}$ only. The world model and actor–critic are optimized on dense rewards, with no access to the sparse task reward and no distillation

    \item \textbf{Sparse:}
    A Dreamer agent trained directly on the sparse task reward $r^{\text{sparse}}$ without any privileged information. The world model predicts sparse rewards and the actor–critic is optimized purely on sparse returns; no HER and no distillation are used.

    \item \textbf{Sparse + HER:}
    Same as \emph{Sparse}, but the replay buffer is wrapped with hindsight experience replay (HER). For each trajectory from the base buffer, the HER wrapper yields the original sequence and $k=4$ relabeled copies: for each copy, a new desired goal is sampled using the ``future'' strategy within a horizon of 5 steps from the same episode, and rewards are recomputed by the task reward function on the achieved-goal and the relabeled desired goal.  Discount and terminal flags are truncated at the first successful step so that relabeled successes do not bootstrap beyond goal achievement.

    \item \textbf{Sparse + D + HER:}
    A sparse-reward student with both world-model distillation and HER. The student world model is regularized toward a frozen dense-reward teacher via latent alignment (posterior and prior), while HER-augmented replay (as above) is used to provide additional successful sparse-reward trajectories.

    \item \textbf{Sparse + D (Ours):}
    Our proposed method: a sparse-reward student with world-model distillation but no HER.  The teacher is trained only on $r^{\text{dense}}$; its world model is then frozen and used as a latent distillation target for the student world model. The student’s actor and critic are trained solely on the sparse reward model, so any improvement over the dense teacher arises from re-optimizing the policy under $r^{\text{sparse}}$ in a teacher-shaped latent space.
    
\end{itemize}

All baselines share the same observations, action space, optimizer, and compute budget.

\subsection{Evaluation Metrics}

For lane-following we evaluate policies using safety, goal-reaching, and lane-keeping metrics derived from CARLA sensors.  
All metrics are computed per episode and then averaged over routes and three random seeds; Tables~\ref{tab:lane_unseen}--\ref{tab:lane_seen_unseen} report mean $\pm$ standard deviation.

\begin{itemize}
\item \textbf{Success [\%].}
Fraction of episodes that reach the final waypoint within the time limit without collisions or terminal infractions (e.g., major lane departure).  
This matches the sparse task success condition used for training.

\item \textbf{Coll./km.}
Number of collision events per kilometre travelled.  
We count all contacts reported by CARLA’s collision sensor and normalize by the total driven distance, so that longer routes do not trivially produce higher counts.

\item \textbf{Lane inv./km.}
Number of lane-invasion events per kilometre travelled.  
Lane invasions are triggered when any wheel crosses a lane marking; we again normalize by total distance to obtain a route-length–independent rate.

\item \textbf{Off-centre [m].}
Mean absolute lateral deviation of the ego vehicle from the reference lane centreline, measured in metres and averaged over all time steps of the episode.  
Lower values indicate better lane keeping.
\end{itemize}

For overtaking we evaluate policies using task-completion, manoeuvre-specific, and safety
metrics derived from CARLA. All metrics are computed per episode and then averaged over
routes and three random seeds; Tables~\ref{tab:overtaking_unseen}--\ref{tab:overtaking_seen_unseen}
report mean $\pm$ standard deviation.

\begin{itemize}
\item \textbf{Success [\%].}
Fraction of episodes that reach the goal region within the time limit without collisions
or terminal infractions. This corresponds to the sparse task success condition for the
overtaking benchmark.

\item \textbf{Overtake [\%].}
Fraction of episodes in which the ego vehicle completes a full overtake: it passes the
slower lead vehicle and returns to its original lane ahead of it without colliding or
timing out.

\item \textbf{Exceed [\%].}
Fraction of episodes in which the ego vehicle exceeds the lead vehicle at any point
(i.e., attains a longitudinal position in front of it), regardless of whether it
returns to the lane and finishes the manoeuvre. This captures partial or unfinished
overtakes.

\item \textbf{Coll./km.}
Number of collision events per kilometre travelled, computed from CARLA’s collision
sensor and normalized by the total driven distance.

\item \textbf{Lane inv./km.}
Number of lane-invasion events per kilometre, where an invasion is triggered whenever a
wheel crosses a lane marking.

\item \textbf{Off-centre [m].}
Mean absolute lateral deviation of the ego vehicle from the reference lane centreline,
in metres, averaged over all time steps of the episode. Lower values indicate better
lane keeping during and after the overtake.
\end{itemize}

\textbf{Generalization.} We report metrics on unseen routes of Town07 and Town04 for lane following and ovetaking respectively. For each metric we report mean $\pm$ standard deviation across seeds. 

\subsection{Reproducibility}\label{sec:repro}
All code is in Python with CARLA~0.9.15 and JAX (using our Dreamer-style implementation).
Training runs on a single NVIDIA RTX~6000 Ada GPU. We will release environment definitions, and configuration files specifying learning rates, latent sizes, sequence lengths, batch sizes, the distillation weight $\beta$, and random seeds.

\section{Results} \label{sec:results}

We evaluate five agents on two CARLA tasks: lane‑following in Town07 and overtaking in Town04. All methods share the same visual observations and action space.

\begin{itemize}
    \item Dense – DreamerV3 trained with dense reward.
    \item Sparse – DreamerV3 trained with sparse task reward only.
    \item Sparse + HER – Sparse reward with Hindsight Experience Replay.
    \item Sparse + D + HER – Sparse + HER with additional world‑model distillation.
    \item Sparse + D (Ours) – Sparse reward with world‑model distillation (no HER).
\end{itemize}

For each setting we report means and standard deviations over three seeds.
Success is the percentage of episodes that reach the goal (lane‑following) or complete a full overtake (overtaking) without collision. Coll./km and Lane inv./km are the number of collisions and lane invasions per kilometre travelled. Off‑centre is the mean lateral deviation from the lane centre in metres.

\subsection{Lane-following results}

\paragraph{Performance on unseen routes.}
Table~\ref{tab:lane_unseen} reports lane-following performance on the unseen Town07 routes.
The dense-reward Dreamer baseline (“Dense”) reaches the goal without collision in
$40.30{\pm}1.96\%$ of episodes, while the sparse-reward baseline (“Sparse”) almost never
succeeds ($1.17{\pm}1.65\%$), despite driving aggressively, with
$22.94{\pm}2.72$ collisions/km and $225.58{\pm}8.14$ lane invasions/km.
Adding HER to the sparse agent (“Sparse + HER”) substantially increases success to
$45.52{\pm}2.05\%$, but at the cost of poorer lane keeping: lane invasions rise to
$168.40{\pm}3.66$ per km and off-centre error remains relatively high
($0.706{\pm}0.009$,m).

Reward‑privileged world‑model distillation without HER (“Sparse + D (Ours)”) achieves the
best overall performance on unseen routes.
Our student reaches the goal in $49.52{\pm}1.40\%$ of episodes, a
$\sim 23\%$ relative improvement over the dense teacher and an $\sim 9\%$ improvement over
the HER-augmented sparse baseline.
At the same time, safety and lane keeping match or improve on the dense teacher:
collisions are slightly reduced ($9.68{\pm}0.39$ vs.\ $10.0{\pm}0.13$ collisions/km),
lane invasions are comparable ($68.97{\pm}40.93$ vs.\ $68.76{\pm}40.71$ per km),
and the mean lateral error is reduced from $0.82{\pm}0.02$,m to
$0.736{\pm}0.014$,m (about $10\%$ lower).
By contrast, combining distillation with HER (“Sparse + D + HER”) fails to learn a
useful policy, with success below $2\%$ and safety metrics close to the unstable
sparse baseline, suggesting an adverse interaction between hindsight relabelling and
the distilled latent dynamics.

\paragraph{Generalization from seen to unseen routes.}
Table~\ref{tab:lane_seen_unseen} summarizes generalization from the training
sub-routes in Town07 (“Seen”) to the disjoint evaluation routes (“Unseen”).
On the seen routes, the dense teacher solves most episodes
($69.51{\pm}4.99\%$ success), while the pure sparse agent again fails to learn
($0\%$ success).
HER makes sparse training competitive with the dense teacher
($66.34{\pm}3.12\%$ success and only $0.48{\pm}0.60$ collisions/km), but with
much poorer lane discipline ($158.94{\pm}2.90$ lane invasions/km and
$0.392{\pm}0.006$,m off‑centre).

Reward‑privileged world‑model distillation leads to both higher success and improved
stability on the training routes.
Our student attains $92.48{\pm}3.83\%$ success—about $33\%$ relative improvement
over the dense teacher and $39\%$ over the HER baseline—while maintaining very low
collision rates ($0.41{\pm}0.53$ collisions/km) and the best lane keeping
($0.071{\pm}0.001$,m off‑centre).
When evaluated on unseen routes, this policy still achieves the highest absolute
success rate ($49.52{\pm}1.40\%$) among all methods, with safety metrics similar to
the dense teacher.
Overall, these results indicate that using dense rewards only to train a teacher world
model, and then distilling its latent dynamics into a sparse‑reward student, yields a
policy that is both safer and more successful than directly optimizing either dense or
sparse rewards (with or without HER) in the lane-following task.

\begin{table*}[t]
  \centering
  \caption{
    Lane-following performance on \textbf{unseen} Town07 routes averaged over three random seeds.
    Success denotes the percentage of episodes that reach the goal without collision.
    \textit{Coll./km} and \textit{Lane inv./km} are the number of collisions and lane invasions per kilometre travelled.
    \textit{Off-centre} is the mean lateral deviation from the lane centre in metres.
  }
  \label{tab:lane_unseen}
  \begin{tabular}{lcccc}
    \toprule
    Method 
                                & Success [\%] 
                                & Coll./km 
                                & Lane inv./km 
                                & Off-centre [m] \\
                                
    \midrule
    Dense                       & $ 40.30 \pm 1.96 $ 
                                & $ 10.0 \pm 0.13 $ 
                                & $ \textbf{68.76} \pm \textbf{40.71} $ 
                                & $ 0.82 \pm 0.02 $ \\
                                
    Sparse                      
                                & $ 1.17 \pm 1.65 $ 
                                & $ 22.94 \pm 2.72 $ 
                                & $ 225.58 \pm 8.14 $ 
                                & $ 0.970 \pm 0.061 $ \\
                                
    Sparse + HER                
                                & $ 45.52 \pm 2.05 $ 
                                & $ 16.87 \pm 1.42 $ 
                                & $ 168.40 \pm 3.66 $ 
                                & $ \textbf{0.706} \pm \textbf{0.009} $ \\
                                
    Sparse + D + HER           
                                & $ 1.68 \pm 1.41 $ 
                                & $ 17.33 \pm 4.07 $ 
                                & $ 188.49 \pm 33.70 $ 
                                & $ 0.836 \pm 0.088 $ \\
                                
    Sparse + D (Ours)  
                                & $ \textbf{49.52} \pm \textbf{1.40} $ 
                                & $ \textbf{9.68} \pm \textbf{0.39} $ 
                                & $ \textbf{68.97} \pm \textbf{40.93} $ 
                                & $ \textbf{0.736} \pm \textbf{0.014} $ \\
    \bottomrule
  \end{tabular}

  \vspace{0.3em}
  \raggedright\footnotesize
  Dense = dense-reward world-model RL; Sparse = sparse reward only;
  HER = Hindsight Experience Replay; D = world-model distillation.
\end{table*}

\begin{table*}[t]
  \centering
  \caption{
    Generalization from \textbf{seen} to \textbf{unseen} lane-following routes in Town07.
    Means $\pm$ standard deviations over three random seeds.
    We report success rates, safety metrics, and lane-keeping performance.
  }
  
  \label{tab:lane_seen_unseen}

  {\setlength{\tabcolsep}{3pt}
  \begin{scriptsize}
  
  \begin{tabular}{lcccccccc}
    \toprule
    & \multicolumn{2}{c}{Success [\%]} &
      \multicolumn{2}{c}{Coll./km} &
      \multicolumn{2}{c}{Lane inv./km} &
      \multicolumn{2}{c}{Off-centre [m]} \\
    \cmidrule(lr){2-3}
    \cmidrule(lr){4-5}
    \cmidrule(lr){6-7}
    \cmidrule(lr){8-9}
    Method
      & Seen & Unseen
      & Seen & Unseen
      & Seen & Unseen
      & Seen & Unseen \\
    \midrule
    Dense
      & $ 69.51 \pm 4.99 $ & $ 40.30 \pm 1.96 $
      & $ 2.61 \pm 3.69 $ & $ 10.0 \pm 0.13 $
      & $ \bm{25.64 \pm 2.42} $ & $ 68.76 \pm 40.71 $
      & $ \bm{0.079 \pm 0.003} $ & $ 0.82 \pm 0.02 $ \\
    Sparse
      & $ 0.00 \pm 0.00 $ & $ 1.17 \pm 1.65 $
      & $ 25.00 \pm 1.66 $ & $ 22.94 \pm 2.72 $
      & $ 248.17 \pm 3.58 $ & $ 225.58 \pm 8.14 $
      & $ 0.293 \pm 0.031 $ & $ 0.970 \pm 0.061 $ \\
    Sparse + HER
      & $ 66.34 \pm 3.12 $ & $ 45.52 \pm 2.05 $
      & $ 0.48 \pm 0.60 $ & $ 16.87 \pm 1.42 $
      & $ 158.94 \pm 2.90 $ & $ 168.4 \pm 3.66 $
      & $ 0.392 \pm 0.006 $ & $ 0.706 \pm 0.009 $ \\
    Sparse + D + HER
      & $ 0.00 \pm 0.00 $ & $ 1.68 \pm 1.41 $
      & $ 26.44 \pm 2.67 $ & $ 17.33 \pm 4.07 $
      & $ 280.26 \pm 25.66 $ & $ 188.49 \pm 33.70 $
      & $ 0.261 \pm 0.021 $ & $ 0.836 \pm 0.088 $ \\
    \textbf{Sparse + D (Ours)}
      & $ \bm{92.48 \pm 3.83} $ & $ \bm{49.52 \pm 1.40} $
      & $ \bm{0.41 \pm 0.33} $ & $ \bm{9.68 \pm 0.39} $
      & $ 27.72 \pm 1.36 $ & $ \bm{68.97 \pm 40.93} $
      & $ \bm{0.071 \pm 0.001} $ & $ \bm{0.736 \pm 0.014} $ \\
    \bottomrule
  \end{tabular}
  \end{scriptsize}
  } 

  \vspace{0.3em}
  \raggedright\footnotesize
  Dense = dense-reward world-model RL; Sparse = sparse reward only;
  HER = Hindsight Experience Replay; D = world-model distillation.
\end{table*}

\subsection{Overtaking results}

\paragraph{Performance on unseen routes.}
Table~V reports performance on the unseen Town04 overtaking routes.
The dense-reward teacher (“Dense”) almost never completes a successful overtake
on unseen routes: success is $0.23{\pm}0.22\%$ and full overtakes occur in only
$0.68{\pm}0.25\%$ of episodes, despite frequently pulling out to pass
($8.25{\pm}0.26\%$ exceed rate). Safety is also poor, with
$28.49{\pm}1.64$ collisions/km and $387.84{\pm}13.10$ lane invasions/km and a large
off‑centre deviation of $0.630{\pm}0.004$,m.

Pure sparse-reward training fails to discover the manoeuvre at all:
“Sparse” has zero success and zero overtakes, although it slightly reduces
collisions to $25.67{\pm}0.55$ collisions/km and improves lane keeping.
Adding HER (“S + HER”) produces more aggressive behaviour—episodes exceed the
lead vehicle in $44.61{\pm}0.60\%$ of trials—but still essentially never finish
the manoeuvre (success $0.34{\pm}0.24\%$), while severely degrading lane keeping
($517.44{\pm}8.40$ lane invasions/km). Combining HER with distillation
(“S + D + HER”) again fails to solve the task.

Reward‑privileged world‑model distillation without HER (“Sparse + D (Ours)”)
is the only variant that achieves substantial overtaking on unseen routes.
Our student attains $6.17{\pm}0.85\%$ success and
$11.03{\pm}0.79\%$ full overtakes, corresponding to roughly a $27\times$ relative
improvement in success over the dense teacher ($6.17/0.23{\approx}27$) and a
$16\times$ improvement in completed overtakes ($11.03/0.68$).
The student is also more willing to attempt overtakes, with an exceed rate of
$19.63{\pm}0.35\%$.
Importantly, these gains do not come at the cost of substantially worse safety:
collisions are comparable to the teacher
($27.31{\pm}0.47$ vs.\ $28.49{\pm}1.64$ collisions/km), lane invasions are
reduced ($304.65{\pm}0.86$ vs.\ $387.84{\pm}13.10$ per km), and off‑centre error
improves from $0.630{\pm}0.004$,m to $0.498{\pm}0.002$,m.

\paragraph{Generalization from seen to unseen routes.}
Table~VI shows results on the training routes (“Seen”) and unseen routes
(“Unseen”).
On the seen routes, the dense teacher achieves high success
($95.29{\pm}1.03\%$) and overtaken rates ($71.24{\pm}0.43\%$) with low collision
rates ($2.93{\pm}0.53$ collisions/km).
However, this behaviour does not generalize: on unseen routes, success and
overtaken rates collapse to $0.23{\pm}0.22\%$ and $0.68{\pm}0.25\%$, respectively,
with large increases in collisions and lane invasions.

Sparse baselines again show either complete failure to learn (“Sparse” and
“S + D + HER”, both with $0\%$ success on seen and unseen routes) or very low
success combined with poor safety (“Sparse + HER”).
These results highlight that, for the overtaking task, sparse reward (with or
without HER) is insufficient to learn the long-horizon manoeuvre from scratch.

Our distillation-based student retains almost all of the teacher’s competence on
the seen routes while substantially improving generalization.
On seen routes, “Sparse + D (Ours)” reaches $93.15{\pm}0.43\%$ success and
$93.28{\pm}0.48\%$ completed overtakes—slightly \emph{higher} overtaken rates than
the dense teacher—with modestly higher collision rates
($6.09{\pm}0.42$ vs.\ $2.93{\pm}0.53$ collisions/km) and slightly better lane
keeping ($117.29{\pm}0.23$ vs.\ $126.53{\pm}1.47$ lane invasions/km).
On unseen routes, it achieves the highest success and overtaken rates among all
methods ($6.17{\pm}0.85\%$ success and $11.03{\pm}0.79\%$ overtakes), while
matching or slightly improving the dense teacher’s safety metrics.

Overall, these results show that reward‑privileged world‑model distillation can
turn a dense‑reward overtaking teacher—which overfits almost completely to the
training routes—into a sparse‑reward student that both preserves high
performance on the training distribution and achieves substantially better
generalization and lane keeping on novel overtaking scenarios.

\begin{table*}[t]
  \centering
  \caption{
    Overtaking performance on \textbf{unseen routes} averaged over three random seeds.
    Success and Overtaken denote the percentage of episodes that reach the goal or complete a full overtake.
    \textit{Coll./km} and \textit{Lane inv./km} are the number of collisions and lane invasions per kilometre travelled.
    \textit{Off-centre} is the mean lateral deviation from the lane centre in metres.
  }
  \label{tab:overtaking_unseen}
  \begin{tabular}{lcccccc}
    \toprule
    Method                                                  & Success [\%] 
                                                            & Overtaken [\%] 
                                                            & Exceed [\%] 
                                                            & Coll./km 
                                                            & Lane inv./km 
                                                            & Off-centre [m] \\
                                                            
    \midrule
    Dense                                    
                                                            & $ 0.23 \pm 0.22 $ 
                                                            & $ 0.68 \pm 0.25 $ 
                                                            & $ 8.25 \pm 0.26 $ 
                                                            & $ 28.49 \pm 1.64 $ 
                                                            & $ 387.84 \pm 13.10 $ 
                                                            & $ 0.630 \pm 0.004 $ \\
                                                            
    Sparse               
                                                            & $ 0.00 \pm 0.00 $ 
                                                            & $ 0.00 \pm 0.00 $ 
                                                            & $ 11.83 \pm 1.95 $ 
                                                            & $ 25.67 \pm 0.55 $ 
                                                            & $ \bm{191.55 \pm 12.34} $ 
                                                            & $ \bm{0.407 \pm 0.003} $ \\
                                                            
    S + HER         
                                                            & $ 0.34 \pm 0.24 $ 
                                                            & $ 0.17 \pm 0.24 $ 
                                                            & $ 44.61 \pm 0.60 $ 
                                                            & $ \bm{20.51 \pm 0.64} $ 
                                                            & $ 517.44 \pm 8.40 $ 
                                                            & $ 0.657 \pm 0.003 $ \\
                                                            
    S + D + HER            
                                                            & $ 0.00 \pm 0.00 $ 
                                                            & $ 0.37 \pm 0.26 $ 
                                                            & $ 34.76 \pm 4.12 $ 
                                                            & $ 40.14 \pm 3.43 $ 
                                                            & $ 277.23 \pm 17.29 $ 
                                                            & $ 0.558 \pm 0.009 $ \\
                                                            
    Sparse + D (Ours)           
                                                            & $ \bm{6.17 \pm 0.85} $ 
                                                            & $ \bm{11.03 \pm 0.79}$ 
                                                            & $ \bm{19.63 \pm 0.35}$ 
                                                            & $ 27.31 \pm 0.47$ 
                                                            & $ 304.65 \pm 0.86$ 
                                                            & $ 0.498 \pm 0.002 $ \\
    \bottomrule
  \end{tabular}
\end{table*}

\begin{table*}[t]
  \centering
  \caption{
    Generalization from \textbf{seen} to \textbf{unseen} overtaking routes.
    Means $\pm$ standard deviations over three seeds.
    Dense = dense-reward world-model RL; Sparse = sparse reward only; 
    HER = Hindsight Experience Replay; D = world-model distillation.
  }
  \label{tab:overtaking_seen_unseen}
  \setlength{\tabcolsep}{3pt}
  \begin{scriptsize}
  \begin{tabular}{@{}p{2.3cm}cccccccc@{}}
    \toprule
    & \multicolumn{2}{c}{Success [\%]} &
      \multicolumn{2}{c}{Overtaken [\%]} &
      \multicolumn{2}{c}{Coll./km} &
      \multicolumn{2}{c}{Lane inv./km} \\
    \cmidrule(lr){2-3}
    \cmidrule(lr){4-5}
    \cmidrule(lr){6-7}
    \cmidrule(lr){8-9}
    Method
                                    & Seen & Unseen
                                    & Seen & Unseen
                                    & Seen & Unseen
                                    & Seen & Unseen \\
    \midrule
    Dense         
                                    & $ \bm{95.29 \pm 1.03} $ & $ 0.23 \pm 0.22 $
                                    & $ 71.24 \pm 0.43 $ & $ 0.68 \pm 0.25 $
                                    & $ \bm{2.93 \pm 0.53} $ & $ 28.49 \pm 1.46 $
                                    & $ 126.33 \pm 1.47 $ & $ 387.84 \pm 13.10 $ \\
                                    
    Sparse
                                    & $ 0.00 \pm 0.00 $ & $ 0.00 \pm 0.00 $
                                    & $ 0.00 \pm 0.00 $ & $ 0.00 \pm 0.00 $
                                    & $ 36.11 \pm 0.47 $ & $ 25.67 \pm 0.55 $
                                    & $ 170.26 \pm 6.33 $ & $ 191.55 \pm 12.34 $ \\
                                    
    Sparse + HER 
                                    & $ 1.06 \pm 0.55 $ & $ 0.34 \pm 0.24 $
                                    & $ 13.30 \pm 1.44 $ & $ 0.17 \pm 0.24 $
                                    & $ 16.98 \pm 1.67 $ & $ 20.15 \pm 0.64 $
                                    & $ 299.55 \pm 1.32 $ & $ 517.44 \pm 8.40 $ \\
                                    
    Sparse + D + HER
                                    & $ 0.00 \pm 0.00 $ & $ 0.00 \pm 0.00 $
                                    & $ 0.00 \pm 0.00 $ & $ 0.37 \pm 0.26 $
                                    & $ 79.88 \pm 0.37 $ & $ 40.14 \pm 3.43 $
                                    & $ 207.94 \pm 7.41 $ & $ 277.23 \pm 17.29 $ \\
                                    
    Sparse + D (Ours)
                                    & $ 93.15 \pm 0.43 $ & $ 6.17 \pm 0.85 $
                                    & $ 93.28 \pm 0.48 $ & $ 11.03 \pm 0.79 $
                                    & $ 6.09 \pm 0.42 $ & $ 27.31 \pm 0.47 $
                                    & $ 117.29 \pm 0.23 $ & $ 304.65 \pm 0.86 $ \\
    \bottomrule
  \end{tabular}
  \end{scriptsize}
\end{table*}

\section{Conclusion}

We introduced \emph{reward-privileged world-model distillation}, where a teacher DreamerV3 agent is trained with a dense simulator reward and only its world model is distilled into a sparse-reward student, without policy or value imitation. Across lane-following and overtaking in CARLA, the distilled students consistently outperform dense-reward teachers and sparse-from-scratch baselines on true task metrics, and are the only agents that reliably generalize overtaking behaviour to unseen routes. Our ablations indicate that aligning latent dynamics, rather than distilling actions or values or adding HER, is key to converting dense-reward competence into sparse-reward performance. This suggests a simple design principle: use privileged dense rewards solely to learn better dynamics models, while optimizing the final policy strictly for deployment-aligned sparse objectives.

\bibliographystyle{IEEEtran}

\bibliography{main}

\end{document}